\documentclass[journal]{IEEEtran}
\pdfoutput=1
\ifCLASSINFOpdf
\else
   \usepackage[dvips]{graphicx}
\fi
\usepackage{url}
\usepackage[dvipsnames]{xcolor}
\usepackage{subfig}
\usepackage{amsfonts}
\usepackage{amsmath}  
\usepackage{caption}
\usepackage{algorithm,algpseudocode}
\newcommand\scalemath[2]{\scalebox{#1}{\mbox{\ensuremath{\displaystyle #2}}}}
\algnewcommand{\algorithmicforeach}{\textbf{for each}}
\algdef{SE}[FOR]{ForEach}{EndForEach}[1]
  {\algorithmicforeach\ #1\ \algorithmicdo}
  {\algorithmicend\ \algorithmicforeach}

\usepackage{lipsum}
\usepackage{graphicx}
\usepackage{units}

\newpage
\begin{document}

\title{Orthogonal Features-based EEG Signal Denoising using Fractionally Compressed AutoEncoder}

\author{Subham Nagar,~\IEEEmembership{}
         Ahlad Kumar,~\IEEEmembership{}
         and M.N.S. Swamy,~\IEEEmembership{Life Fellow,~IEEE}

\thanks{Subham Nagar is currently pursuing his Master's degree in Information and Communication Technology from DA-IICT, Gandhinagar, Gujarat (e-mail: subhamnagar@gmail.com)}
\thanks{Dr. Ahlad Kumar is an Assistant Professor  at DA-IICT. (e-mail: ahlad\_kumar@daiict.ac.in). URL:https://www.daiict.ac.in/profile/ahlad-kumar/}
\thanks{Dr. M.N.S Swamy is working as a Research Professor at Concordia University, Canada. (e-mail: swamy@ece.concordia.ca)}}

\maketitle

\begin{abstract}
A fractional-based compressed auto-encoder architecture has been introduced to solve the problem of denoising electroencephalogram (EEG) signals. The architecture makes use of  fractional calculus to calculate the gradients during the back-propagation process, as a result of which a new hyper-parameter in the form of fractional order $\alpha$ has been introduced which can be tuned to get the best denoising performance. Additionally, to avoid substantial use of memory resources, the model makes use of orthogonal features in the form of Tchebichef moments as input. The orthogonal features have been used in achieving compression at the input stage. Considering the growing use of low energy devices, compression of neural networks becomes imperative. Here, the auto-encoder's weights are compressed using the randomized singular value decomposition (RSVD) algorithm during training while evaluation is performed using various compression ratios. The experimental results show that the proposed fractionally compressed architecture provides improved denoising results on the standard datasets when compared with the existing methods.
\end{abstract}

\begin{IEEEkeywords}
EEG signal denoising, Auto-encoder, Fractional calculus, Orthogonal moments, Compression.
\end{IEEEkeywords}

\IEEEpeerreviewmaketitle

\section{Introduction}

\IEEEPARstart{T}{he} problem of denoising is an important challenge in the field of signal processing. Signals often get corrupted during their storage or transmission, in most cases with Gaussian noise. Hence, it becomes important to restore the original signal from the noisy observations. Various methods have been proposed in order to address this problem, such as adaptive filtering \cite{article1} , principal component analysis \cite{article2} and wavelet transform \cite{article3}. However, a common drawback of all of these methods is the need for analytical calculation and high computation.

Vincent et al. \cite{article4} proposed a machine learning model to detect noise in order to classify unacceptable EEG signals. Li et al. \cite{article5} listed the applications of deep learning architectures in denoising signals by giving an example of a denoising auto-encoder (DAE). The results achieved by DAE has outperformed the conventional methods of denoising signals as can be seen in \cite{article6,article7}. The limitation of using deep learning techniques lies in the fact that large number of weights can lead to a substantial use of memory resources. 

Over the years, several methods have been proposed to address the problem of redundancy in weights of the neural networks. Han et al. \cite{article8} proposed compression by pruning, which uses a hard thresholding technique to remove least important weights from the neural network. Several methods have used the concept of low-rank approximation-based compression, which saves the storage and simultaneously reduced the time complexity during training and testing phase \cite{article9, article10}. 

A method of combining Discrete Cosine Transform (DCT) and deep learning architecture for achieving compression has been proposed in \cite{article11}. Here, the DCT coefficients of input images have been used to represent its prominent features, which were then fed into the neural networks for doing specific tasks. The use of block-wise DCT, when performed on a noisy image, reduces the signal redundancy as a result of which neural networks can learn the important features of image in a better way \cite{article12}. In this paper, first time Tchebichef moments \cite{article16} are used as feature vectors in the proposed deep learning architecture. 

The application of fractional calculus in neural networks has gained prominence. It yields competitive results when compared with integer order deep neural networks \cite{article13}. Fractional calculus has also been extensively investigated in the area of image denoising \cite{df1,df2} and texture enhancement \cite{df3}. This has motivated to explore its potential strength on a deep auto-encoders with the objective of denoising EEG signals. 

In most of the matrix decomposition algorithms used for compressing the weights in the neural network, the randomized singular value decomposition (RSVD) approach has not yet been explored. The concept of calculating rank approximation of matrices using RSVD was introduced in \cite{article14} and has been found to be computationally efficient compared to SVD. In this paper, we apply the RSVD algorithm for compressing the weights in the neural network.

This paper is organized as follows. In Section \ref{sec1} the concepts of fractional calculus, Tchebichef moments and RSVD are introduced along with their respective roles played in implementing the proposed model. Section \ref{sec2} discusses the details of the proposed architecture; Section \ref{sec3} discusses the datasets taken and the metrics used for comparative analysis along with the training and testing results; finally, Section \ref{sec4} concludes this work.

\section{Preliminaries} \label{sec1}
\subsection{Fractional Calculus}
Many definitions have been proposed for fractional order derivatives unlike the integer order derivatives, where we have a unified expression. There are three most commonly used fractional order derivatives, i.e., namely, Grunwald Letnikov (G-L), Riemann-Liouville (R-L) and Caputo derivatives \cite{article22}. The Caputo fractional derivative of a function $f(x)$ with order $\alpha$ is defined as follows:
\begin{equation}
    ^{C}D^{\alpha}f(x)= \frac{1}{\Gamma(n-\alpha)}
    \int_{a}^x \frac{f^{(n)}(y)}{(x-y)^{(1+\alpha-n)}}dy
\end{equation} 
where $n-1 < \alpha < n$. \\
For simplicity, $D^{\alpha}$ defined as the Caputo fractional derivative is employed. It can be observed that $D^{\alpha}$ is consistent with integer order derivatives given the fact that $D^{\alpha}f(x)$ is $0$ for a constant function. Therefore, it is generally used for most of the signal and image processing applications.

\subsection{Tchebichef moment}
The Tchebichef moments of order $m$ for a signal $x(n)$ of $N$ samples is given by \cite{article16}:
\begin{equation}
    T\textsubscript{m}(x) = \sum_{x=0}^{N-1} t\textsubscript{m}(x;N) x(n)
\label{}
\end{equation}
with $m,n = 0, 1 , 2 ..... N-1$. For simplicity, $t\textsubscript{m}(x)$ has been used to represent $t\textsubscript{m}(x;N)$ which is the orthonormal Tchebichef polynomials given by 
\begin{equation}
    t\textsubscript{m}(x) = \alpha\textsubscript{1} (2x+1-N) t\textsubscript{m-1}(x) + \alpha\textsubscript{2} t\textsubscript{m-2}(x)
\label{}
\end{equation}
where
\begin{equation}
    \alpha\textsubscript{1} = \frac{1}{n}\sqrt{\frac{4(n^{2}-1)}{N^{2}-n^{2}}}
\label{}
\end{equation}
\begin{equation}
    \alpha\textsubscript{2} = \frac{1-n}{n}\sqrt{\frac{2n+1}{2n+3}}\sqrt{\frac{N^{2}-(n-1)^{2}}{N^{2}-n^{2}}}
\label{}
\end{equation}
The initial conditions for the recurrence relations are 
\begin{equation}
    t\textsubscript{0}(x) = \frac{1}{\sqrt{n}}
\label{}
\end{equation}
and
\begin{equation}
    t\textsubscript{1}(x) = (2x+1-N)\sqrt{\frac{3}{N(N^{2}-1)}}
\label{}
\end{equation}
The set of Tchebichef moments upto order $p$ in matrix form is given as
\begin{equation}
    T_{p}(\textbf{X}) = \textbf{X}\textbf{Q}^{T}
\label{c1p}
\end{equation}
where $\textbf{X} = [x(0),\; x(1),\; x(2), \ldots,\; x(N-1)]$ and 
\begin{equation}
\textbf{Q} = \begin{bmatrix} 
    t\textsubscript{0}(0) & \dots  & t\textsubscript{0}(N-1) \\
    \vdots & \ddots & \vdots\\
    t\textsubscript{p-1}(0) & \dots  & t\textsubscript{p-1}(N-1)
    \end{bmatrix}
\label{}
\end{equation}
Here, $\textbf{Q}$ is the Tchebichef polynomial matrix upto order $p$, with $n = 1,2, \ldots, N$ and $N$ is the number of samples.
The original one-dimensional signal $\textbf{X}$ can be reconstructed  from the set of Tchebichef moments using the following equation
\begin{equation}
    \textbf{X} = T_{p}(\textbf{X})\textbf{Q}
\label{c10}
\end{equation}
\subsection{RSVD}
The rank approximation of weight matrices used in the proposed architecture have been obtained using the RSVD technique which decomposes the original matrix 
$ A \in  \mathbb{R}^{n\times m} $ into a smaller randomized subspace
$B \in  \mathbb{R}^{c \times m}$ where $c < n$.\\
Let $O \in  \mathbb{R} ^{m \times (r+p)}$ be a normally distributed random matrix, where $r$ is the rank to be approximated, $p$ is the subspace iterations and $r+p < n$. We define $Q\textsubscript{i}$ as the orthogonal basis after $i$ iterations, where $ i = 1,2,3 \ldots, p$. The value of $Q\textsubscript{0}$ is set using the following equation
\begin{equation}
    Q\textsubscript{0} = AO
\label{}
\end{equation}
The recurrence relation for calculating the orthogonal basis $Q\textsubscript{i}$ is given by
\begin{equation}
   G\textsubscript{i} = qr(A\textsuperscript{T}Q\textsubscript{i-1})
\label{c1}
\end{equation}
\begin{equation}
Q\textsubscript{i}  = qr(AG\textsubscript{i})
\label{}
\end{equation}
where $qr()$ is the function for the QR decomposition operation which factorizes a matrix into an orthogonal matrix and an upper triangular matrix. Here, we just take the orthogonal matrix part. The above recurrence equations are applied over $p$ iterations after which we get the value $Q_p$.\\
Now the condensed matrix $B$ can be calculated as
\begin{equation}
   B=Q\textsubscript{p}\textsuperscript{T}A
\label{}
\end{equation}
The SVD decomposition of this condensed matrix is
\begin{equation}
   [U_{r} , S_{r}, V_{r}] = \text{SVD}(B)
\label{}
\end{equation}
\begin{equation}
U\textsubscript{r} = Q\textsubscript{p}U\textsubscript{r}
\label{c1}
\end{equation}
where, $U_{r}\in \mathbb{R} ^ {n\times r} $, $V_{r}\in \mathbb{R} ^{m\times r}$ are the matrices with orthonormal columns and $S_{r}\in \mathbb{R}^{r\times r}$  is a diagonal matrix.

\begin{figure*}
    \centering
    \includegraphics[width=7.5in,height=3.5in]{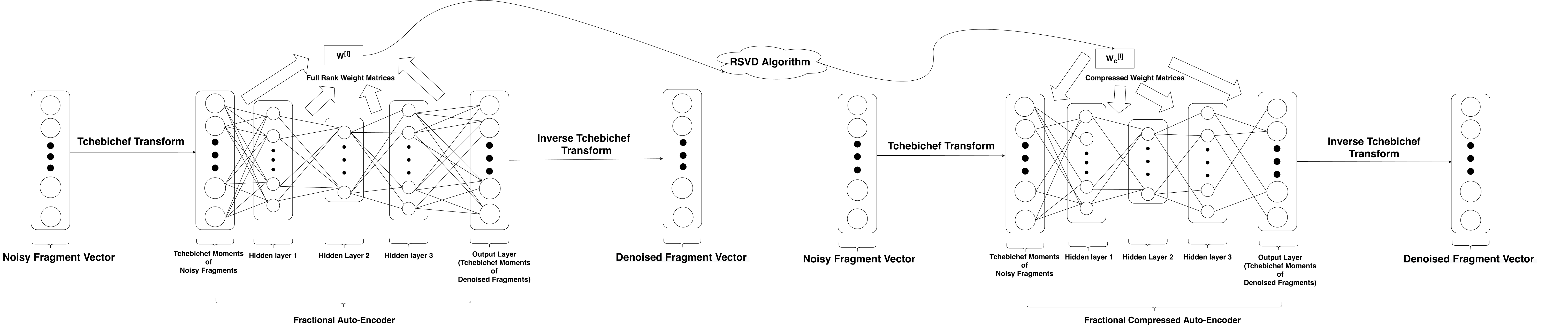}
    \caption{Architecture of the proposed Auto-Encoder}
    \label{fig1}
\end{figure*}

\section{The proposed model and algorithm} \label{sec2}
Let $x(n)$ be an EEG signal with $n=1,2, \ldots, N$. The relationship between the noisy signal $y(n)$ and the original signal $x(n)$ corrupted by noise is given as follows:
\begin{equation}
   y(n) = x(n)+ \zeta(n)
\label{c1}
\end{equation}
where $\zeta$ is the Gaussian noise having mean zero and variance $\sigma^{2}$. The main contribution of this paper is to recover the original signal $x(n)$ given the noisy version $y(n)$. The proposed fractional auto-encoder and its compressed architecture used for this purpose are shown in Fig \ref{fig1}. The fractional auto-encoder is utilized for training, which uses fractional back-propagation technique. For the evaluation of the model on datasets, the compressed form of this network (Fractional Compressed Auto-Encoder) is used. The compressed version takes trained weights and finds the corresponding low rank approximation using RSVD compression. Both these architectures use orthogonal features as input, which are obtained using Tchebichef transform of EEG signals. It has been shown earlier in \cite{8611418} that Tchebichef moments help in denoising the images. Motivated by this work, one dimensional version of the Tchebichef orthogonal moments is applied on noisy input signals. The transformed signal in the orthogonal space acts as an input feature to the architecture. This helps in not only representing the original signal in compressed form, but also helps in addressing the problem of signal denoising. The weights present in the architecture are also compressed using RSVD algorithm during training and testing phase. The details of this will be discussed in the experimental study section.\\
The auto-encoder architecture shown in Fig. \ref{fig1} has five layers namely, one input layer, three hidden layers and one output layer. In order to feed the orthogonal features to the auto-encoder, we first transform the input signal into Tchebichef moment (orthogonal) space using Eq. \ref{c1p}, where $\textbf{X}$ is the input signal and $p$ is the Tchebichef polynomial upto order $N$. The output layer of the architecture reconstructs the Tchebichef moments of denoised signal $x(n)$ given the Tchebichef moments of noisy EEG signal $y(n)$ as input to the network. In the task of signal denoising, both the input layer and the output layer have $N$ neurons. The first and the penultimate hidden layers have $N_h$ number of neurons, while the middle hidden layer has $N_e$ number of neurons, where $N_e < N_h < N$. Moreover, min-max normalization technique is used to reduce the effect of the outliers \cite{article18}. Each layer of the architecture uses a linear activation function except for the output layer, which uses $sigmoid$ function. The reason for selecting this activation is that the Tchebichef moments of the original signal $T_{N}(x)$ lie in the range of [0,1], given the fact that each input is normalized in that range. Here, $W\textsuperscript{[$l$]}$ is used to denote the weight matrix storing the weights of the connections from layer $l$ to layer $(l-1)$ and $B\textsuperscript{[$l$]}$ to denote the bias values with respect to layer $l$.\\
In order to denoise the EEG signals, the next step involved is to train the auto-encoder. The details of the training process is depicted in the $Algorithm \; 1$. The values of the pre-activated and activated neurons in the hidden layers are denoted by $Z\textsuperscript{[$l$]}$ and $A\textsuperscript{[$l$]} \in \mathbb{R}\textsuperscript{N\textsubscript{$l$}}$, where $N\textsubscript{$l$}$ is the number of nodes in the $l\textsuperscript{th}$ hidden layer. The forward propagation is performed using the following equations
\begin{equation}
    Z\textsuperscript{[$l$]} = W\textsuperscript{[$l$]}A\textsuperscript{[$l-1$]} + B\textsuperscript{[$l$]}
\label{c2}
\end{equation}
\begin{equation}
    A\textsuperscript{[$l$]} = g\textsuperscript{[$l$]} (Z\textsuperscript{[$l$]})
\label{c3}
\end{equation}
for $l$ = 1,2,3,4. Here, $A\textsuperscript{[0]}$ represents the input features, which are basically the  Tchebichef moments of the noisy signal $T_{N}(y)$ fed to the auto-encoder and g\textsuperscript{$[l]$} corresponds to the activation function in the $l\textsuperscript{th}$ hidden layer.
\begin{algorithm}
  \caption{Training of the proposed architecture}
  \begin{algorithmic}[1]
    \State \textbf{Input: } Trainable parameters $\{W\textsuperscript{[$l$]} , B\textsuperscript{[$l$]}\}_{l=1}^L$ 
    \ForEach{$i$ in $epochs$}
    \State shuffle training data
    \ForEach{$k$ in $iterations$}
    \State Forward Propagation  \algorithmiccomment using Eqs. (\ref{c2})- (\ref{c3})
    \State Backward Propagation \algorithmiccomment using $algorithm 2$
    \State Update Parameters \algorithmiccomment  using Eqs.(24)-(25);
    \EndForEach
    \State \textbf{if} (convergence criterion met)
    \ForEach{$l$ in ${1,2,..,L}$}%
    \State $r = rank(W^{[l]})$
    \State $W_{c}^{[l]}\gets$ $RSVD\_Opt\_Compression$($W\textsuperscript{[$l$]}$ , $r$)
    \EndForEach
    \State \textbf{end if}
    \EndForEach
    \State \textbf{return} $A\textsuperscript{[L]}$
  \end{algorithmic}
\end{algorithm}

The output unit $A\textsuperscript{[4]}$ contains the final values of the auto-encoder, which are the Tchebichef moments of the estimated denoised signal  $(\hat T_{N}(x))$. The proposed regularized  cost function $J$ used for the auto-encoder is given by
\begin{equation}
\scalemath{0.9}{
   J(W,B) = \frac{1}{2M}  \sum_{i=1}^{M}\sum_{l=1}^{4}\left(T_{N}^{[l](i)}(x) - \hat T_{N}^{[l](i)} (x)\right)^{2} +\frac{\lambda}{2}\left\lVert (W^{[l]})^{2} \right\rVert
\label{c4}}
\end{equation}
where $M$ is the total number of training samples and $\lambda$ is the regularization parameter, whose value is taken as $10^{-6}$.The trainable parameters $(W,B)$ present in the cost function are optimized via fractional back-propagation technique in which the fractional derivatives of the weights are taken rather than integer order derivatives. The details of this are been given in $Algorithm \; 2$. The fractional-order gradient descent equation for both the trainable parameters are given by
\begin{equation}
   W\textsuperscript{[$l$]}\gets W\textsuperscript{[$l$]} - \eta D^\alpha W\textsuperscript{[$l$]}
\label{c5}
\end{equation}
\begin{equation}
   B\textsuperscript{[$l$]}\gets B\textsuperscript{[$l$]} - \eta dB\textsuperscript{[$l$]}
\label{c6}
\end{equation}
where $D^{\alpha} W^{[l]} = \dfrac{\partial^{\alpha} J(W,B)}{\partial^{\alpha} W^{[l]}}$, $d{B^{[l]}} = \dfrac{\partial J(W,B)}{\partial B^{[l]}}$ and $\eta$ is the learning rate.
\begin{algorithm}
  \caption{ Fractional Backward Propagation}
  \begin{algorithmic}[1]
    \State \textbf{procedure} Backward Propagation:
    \State $dZ\textsuperscript{[L]}\gets (A\textsuperscript{[L]} - Y) * g\textsuperscript{[L]}'(Z\textsuperscript{[L]})$ ; 
    \ForEach{$l$ in ${L , L-1  ,... , 2}$}%
      \State $D^\alpha W\textsuperscript{[$l$]}\gets (d Z\textsuperscript{[$l$]}(A\textsuperscript{[$l-1$]})\textsuperscript{T}) \cdot \frac{(W\textsuperscript{[$l$]})\textsuperscript{(1-$\alpha$)}}{\Gamma(2-\alpha)} $ \algorithmiccomment  from $(2)$
      \State $D^\alpha W\textsuperscript{[$l$]}\gets D^\alpha W\textsuperscript{[$l$]}+\lambda (\frac{W\textsuperscript{[$l$]}\textsuperscript{(2-$\alpha$)}}{\Gamma (3-\alpha)}) $ \algorithmiccomment  from $(2)$ 
      \State $dB\textsuperscript{[$l$]}\gets dZ\textsuperscript{[$l$]}$ 
      \State $dA\textsuperscript{[$l-1$]}\gets (W\textsuperscript{[$l$]})\textsuperscript{T}d Z\textsuperscript{[$l$]}$
      \State $dZ\textsuperscript{[$l-1$]}\gets dA\textsuperscript{[$l-1$]} \cdot  g\textsuperscript{[$l-1$]}'(Z\textsuperscript{[$l-1$]})$ 
    \EndForEach
    \State $D^\alpha W\textsuperscript{[1]}\gets (d Z\textsuperscript{[1]}(A\textsuperscript{[0]})\textsuperscript{T}) \cdot \frac{(W\textsuperscript{[1]})\textsuperscript{(1-$\alpha$)}}{\Gamma(2-\alpha)}$  \algorithmiccomment  from $(2)$
      \State $D^\alpha W\textsuperscript{[1]}\gets D^\alpha W\textsuperscript{[1]}+\lambda (\frac{W\textsuperscript{[1]}\textsuperscript{(2-$\alpha$)}}{\Gamma(3-\alpha)}) $ \algorithmiccomment  from $(2)$
   \State $dB\textsuperscript{[1]}\gets dZ\textsuperscript{[1]}$
   \State \textbf{end procedure}
  \end{algorithmic}
\end{algorithm}
\begin{algorithm}
  \caption{RSVD compression for obtaining compressed Weight matrices}
  \begin{algorithmic}[1]
  \State \textbf{procedure} $RSVD\_Opt\_Compression$($W^{[l]}$ , $r$):
 \State calculate [U\textsubscript{$r$} , S\textsubscript{$r$} , V\textsubscript{$r$}] = RSVD ($W^{[l]}$ , $r$) \algorithmiccomment from Eqs. $(12)$-$(17)$ 
 \State $opt$$\gets$ $check\_optimized\_rank$ ($S\textsubscript{$r$}$)
 \State $W_{c}^{[l]}\gets U\textsubscript{opt}S\textsubscript{opt}V_{opt}^{T}$ 
 \State \textbf{return} $W\textsubscript{c}$
 \State \textbf{end procedure}
  \end{algorithmic}
\end{algorithm}

The training procedure also includes the RSVD compression, which finds the optimized rank of the weight matrices $W^{[l]}$ approximated at every 200 epochs. The procedure associated with this process is denoted by $RSVD\_Opt\_Compression$ given in $Algorithm \; 3$, which takes the weight matrix and its respective rank $(r)$ as the input parameters. The optimized rank is selected as the value, where 90\% of the variance is covered for singular vectors obtained, using RSVD decomposition. This is done using $check\_optimized\_rank$ function mentioned in $Algorithm \; 3$. The approximation of the weight matrices $W^{[l]}$ are, then, calculated using the optimized rank. This helps in improving the performance of the auto-encoder, when the weights are compressed with respect to various rates. The final evaluation of our proposed model is carried out in $Algorithm \; 4$. The main objective is to evaluate the denoising performance using the compressed auto-encoder. Therefore, RSVD compression on the trained weight matrices is carried out using a specified compression rate ($C\textsubscript{R}$). The range of $C\textsubscript{R}$ is taken to be between $0.5$ to $0.95$. The compression is performed on the weight matrices $W^{[l]}$ by finding its approximated matrix $W\textsubscript{c}\textsuperscript{[$l$]}$,  which has a reduced rank $r$ calculated in $Algorithm \; 4$. Once the compressed matrices $(W_{c}^{[l]},B^{[l]})$ have been obtained, the testing of the proposed architecture is carried out using the new noisy EEG observations as input, and the denoised output performance is evaluated using the metrics discussed in the experimental section.

\begin{algorithm}
  \caption{Evaluating performance using compressed weight matrices}
  \begin{algorithmic}[1]
    \State \textbf{Input: } Trained parameters $\{W\textsuperscript{[$l$]} , B\textsuperscript{[$l$]}\}_{l=1}^L$ \algorithmiccomment  from $algorithm 1$
    \State Initialize compression rate $C\textsubscript{R}$
    \ForEach{$l \in {1 , 2  ,... , L}$}%
     \State $r = (1-C\textsubscript{R}) * rank(W^{[l]})$ 
 \State [U\textsubscript{$r$} , S\textsubscript{$r$} , V\textsubscript{$r$}] = RSVD ($W\textsuperscript{[$l$]}$, $r$) \algorithmiccomment from Eqs.$(12)$-$(17)$ 
 \State $W_{c}^{[l]} \gets U\textsubscript{$r$}S\textsubscript{$r$}V_{r}^{T}$
    \State \textbf{Forward Propagation } 
    \State $Z\textsuperscript{[$l$]}\gets W_{c}^{[l]} A\textsuperscript{[$l-1$]} + B\textsuperscript{[$l$]}$
    \State $A\textsuperscript{[$l$]}\gets g\textsuperscript{[$l$]} (Z\textsuperscript{[$l$]})$
    \EndForEach
    \State \textbf{return} $A\textsuperscript{[L]}$
  \end{algorithmic}
\end{algorithm}

\section{Experimental Study } \label{sec3}
\subsection{Datasets and Metrics} \label{A}
To test the effectiveness of the proposed approach, two standard EEG signal datasets namely, the Keirn's EEG dataset \cite{article20} and Motor Movement/Imagery \cite{article21} are used. In the Keirn's EEG dataset, six subjects contributed in recording EEG signals from six electrodes namely, C3, C4, P3, P4, O1, and O2. Each volunteer participates in five mental tasks during the recording, with each task repeated three times for a couple of sessions. In each session, the subject's recording has been done with both of the positions: rest eyes closed and rest eyes open. Every task has a recording period of 10 seconds each, with a sampling rate of 250 Hz per second. The age range of participants is between 21 and 48.\\ 
The Motor Movement/Imagery has the contribution of 109 healthy subjects who recorded the EEG signal using BCI2000 system, which is a brain computer interface software. The signals have been recorded using 64 electrodes, each having a sampling rate of 250 Hz per second. Several tasks like imagining and simulating a given action, like opening and closing of eyes have been performed by the volunteers.
The evaluation of the proposed approach is carried out based on two performance metrics, which are the Percentage Root Mean Square Difference ($PRD$) and Signal-to-Noise Ratio ($SNR$)  given by 
\begin{equation}
   PRD = 100 * \sqrt{\frac{\sum\limits_{n=1}^{N}[x(n) - \hat x(n)]^{2}}{\sum\limits_{n=1}^{N}[x(n)]^{2}}}
\label{c7}
\end{equation}

\begin{equation}
   SNR = 10 * \log_{10} \frac{\sum\limits_{n=1}^{N}[x(n)]^{2}}{\sum\limits_{n=1}^{N}[x(n) - \hat x(n)]^{2}}
\label{c8}
\end{equation}
where $x(n)$ is the original signal, $\hat x(n)$ is the reconstructed signal and $N$ is the total number of samples.

\subsection{Training Details}
The Keirn dataset contains 2100 signals, each of sample length $(N)$ $2500$. The Gaussian noise of standard deviation ($\sigma$) 15 is added to the original signals ( see Eq. \ref{c1}). We first split each of the noisy signals into fragments of 250 samples such that no overlapping occurs. The Tchebichef moment (orthogonal feature) is calculated for each of the noisy signal vector using Eq. (\ref{c1p}). The fractional auto-encoder is trained using these transformed noisy signals. Once the auto-encoder has been trained, we reconstruct all the fragmented denoised Tchebichef moments $\hat T_{N}(x)$ for the corresponding noisy signal $y(n)$ and concatenate them to obtain the final denoised signal $\hat x(n)$ using Eq. (\ref{c10}). The quality of the restored signal is evaluated using the performance metrics given by Eqs. (\ref{c7})-(\ref{c8}). 

Normalization on the orthogonal features is carried out to scale the values in the range of $(0,1)$. The training of the fractional auto-encoder is performed by splitting the normalized data into $70\%$ training and $30\%$testing set. For the training process, the epochs, batch size and learning rate $(\eta)$ are kept as $3000$, $256$ and $0.01$, respectively. As mentioned in $Algorithm \; 3$, the optimized RSVD compression is performed on the weights after every 200 epochs. The training and testing loss are monitored to check whether the model is over-fitting or not. It has been found that the above parameters selected for training are optimal for the proposed architecture. The same process is implemented for the Motor-EEG dataset, the only change being that the epochs and batch size are set to $500$ and $1024$, respectively. Here, the weight compression is performed after every 50 epochs. All the experiments were performed on NVIDIA K80 GPU.

 \begin{table}[htbp]
\renewcommand{\arraystretch}{1.3}
\captionsetup{justification=centering, labelsep=newline,font=footnotesize,labelfont=normalsize}
\caption{ Comparison of the proposed method with the existing methods for Keirn dataset}
\label{tab:1a}
\centering
\begin{tabular}{c|c|c}
\hline
\multicolumn{1}{c}{\textbf{Method}} &  \multicolumn{1}{|c|}{\textbf{SNR}} & \multicolumn{1}{|c}{\textbf{PRD}} \\
\hline
Kumari Method \cite{KUMARI} & 0.52 &   94.10 \\
Al-Quazzaz Method \cite{quazzaz} & 0.66 & 92.66\\
FPA-WT \cite{fpa} & 2.20 &  78.76 \\
Proposed Method ($\alpha = 1$) & 0.49 &  94.54 \\
\textbf{Proposed Method ($\alpha = 1.7$)} &  \textbf{2.60} &   \textbf{75.47} \\
\hline
\end{tabular}
\end{table}

\begin{table}[htbp]
\renewcommand{\arraystretch}{1.3}
\captionsetup{justification=centering, labelsep=newline,font=footnotesize,labelfont=normalsize}
\caption{Comparison of the proposed method with the existing methods for Motor EEG dataset}
\label{tab:2a}
\centering
\begin{tabular}{c|c|c}
\hline
\multicolumn{1}{c}{\textbf{Method}} &  \multicolumn{1}{|c|}{\textbf{SNR}} & \multicolumn{1}{|c}{\textbf{PRD}} \\
\hline
FPA-WT \cite{fpa} & 0.46 &  98.75 \\
Proposed Method ($\alpha = 1$) & 5.32 &   55.78 \\
\textbf{Proposed Method ($\alpha = 1.6$)} &  \textbf{6.41} &   \textbf{49.51} \\
\hline
\end{tabular}
\end{table}
\subsection{Signal Denoising}
\begin{figure*}[!ht]
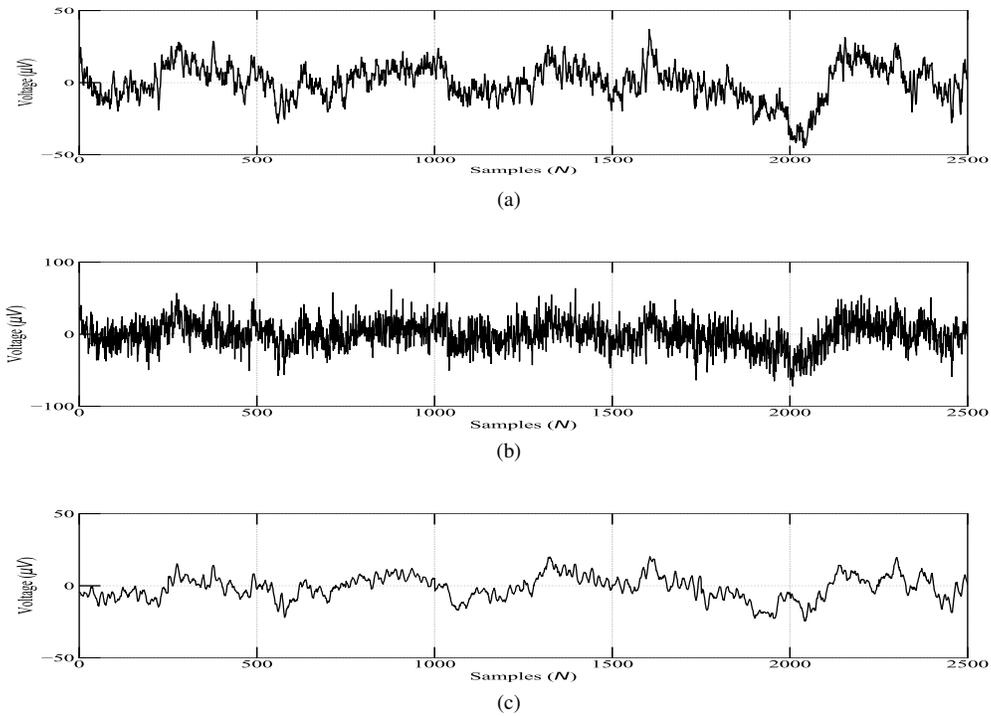

\centering
\subfloat[\label{fig:test1}]
  {\includegraphics[width=6.0in,height=1.0in]{Figure_2.pdf}} \hfill
\subfloat[\label{fig:test2}]
  {\includegraphics[width=6.0in,height=1.0in]{Figure_3.pdf}} \hfill
\subfloat[\label{fig:test3}]
  {\includegraphics[width=6.0in,height=1.0in]{Figure_4.pdf}}\hfill
\caption{EEG signal denoising using the proposed architecture on Keirn Dataset: (a) original signal (b) noisy signal and (c) denoised signal }
\label{g1}
\end{figure*}

In this section, the performance of the proposed fractional auto-encoder in the area of signal denoising is demonstrated. 
For this purpose, simulations have been carried on two datasets, namely, Keirn and  Motor Movement/Imagery, and the details are discussed in Section \ref{A}. The proposed deep learning framework is developed for restoring the EEG signals degraded due to the Gaussian noise. Hence, it becomes essential to compare the quality of the restored signals with that of the other state-of-the-art algorithms. For
this purpose, in the case of Gaussian noise, we consider FPA-WT \cite{fpa}, Al-Quazzaz-Method \cite{quazzaz} and Kumari Method \cite{KUMARI} that use wavelet transform for solving the problem of EEG signal denoising. \\
The performance of the proposed architecture on sample EEG signals taken from the two datasets is visualized in Figs. \ref{g1} and \ref{g2}, respectively. The signals shown in Figs. \ref{g1}(a) and \ref{g2}(d) represents the original EEG signals; the corresponding noisy signals are shown in Figs. \ref{g1}(b) and \ref{g2}(e). Here, the noisy signals have been obtained by corrupting the original signal with Gaussian noise of standard deviation ($\sigma$) of 15. It can be observed from Figs. \ref{g1}(c) and \ref{g2}(f) that the proposed architecture are very effective in denoising the noisy signals. The $SNR$ and $PRD$ values for the denoised signals shown in Fig. \ref{g1}(c) are 3.75 and 64.93, respectively and that of Fig. \ref{g1}(f), the corresponding values are 9.00 and 35.47.

Table \ref{tab:1a} provides a comparative analysis of the proposed architecture with the existing methods for the Keirn dataset.The performance of the various methods has been evaluated in terms of the average $SNR$ and $PRD$ values. It can be observed that the proposed model provides better performance over the existing methods in terms of these performance metrics. The proposed architecture has been evaluated for different values of $\alpha$. It can be noted that for $\alpha = 1$, the architecture works like a conventional auto-encoder. However, for any other value of $\alpha$ between 1 and 2, it works as a fractional auto-encoder. It is observed that there is a boost in the denoising  performance for $\alpha = 1.7$. Similar analysis has been performed on the Motor EEG dataset with varying fractional orders. The average $SNR$ and $PRD$ values are shown in Table \ref{tab:2a}. It can be observed from the values that even in this case, the proposed model outperforms the existing method \cite{fpa} for $\alpha = 1.6$.  
\begin{figure*}[!ht]
\centering
\subfloat[\label{fig:test4}]
  {\includegraphics[width=6.0in,height=1.0in]{Figure_5.pdf}} \hfill
\subfloat[\label{fig:test5}]
  {\includegraphics[width=6.0in,height=1.0in]{Figure_6.pdf}} \hfill
\subfloat[\label{fig:test6}]
  {\includegraphics[width=6.0in,height=1.0in]{Figure_7.pdf}}\hfill
\caption{EEG signal denoising using the proposed architecture on Motor dataset: (a) original signal (b) noisy signal and (c) denoised signal }
\label{g2}
\end{figure*}
\begin{figure}[!ht]
\centering
\subfloat[\label{test2}]
  {\includegraphics[width=3.5in,height=1.8in]{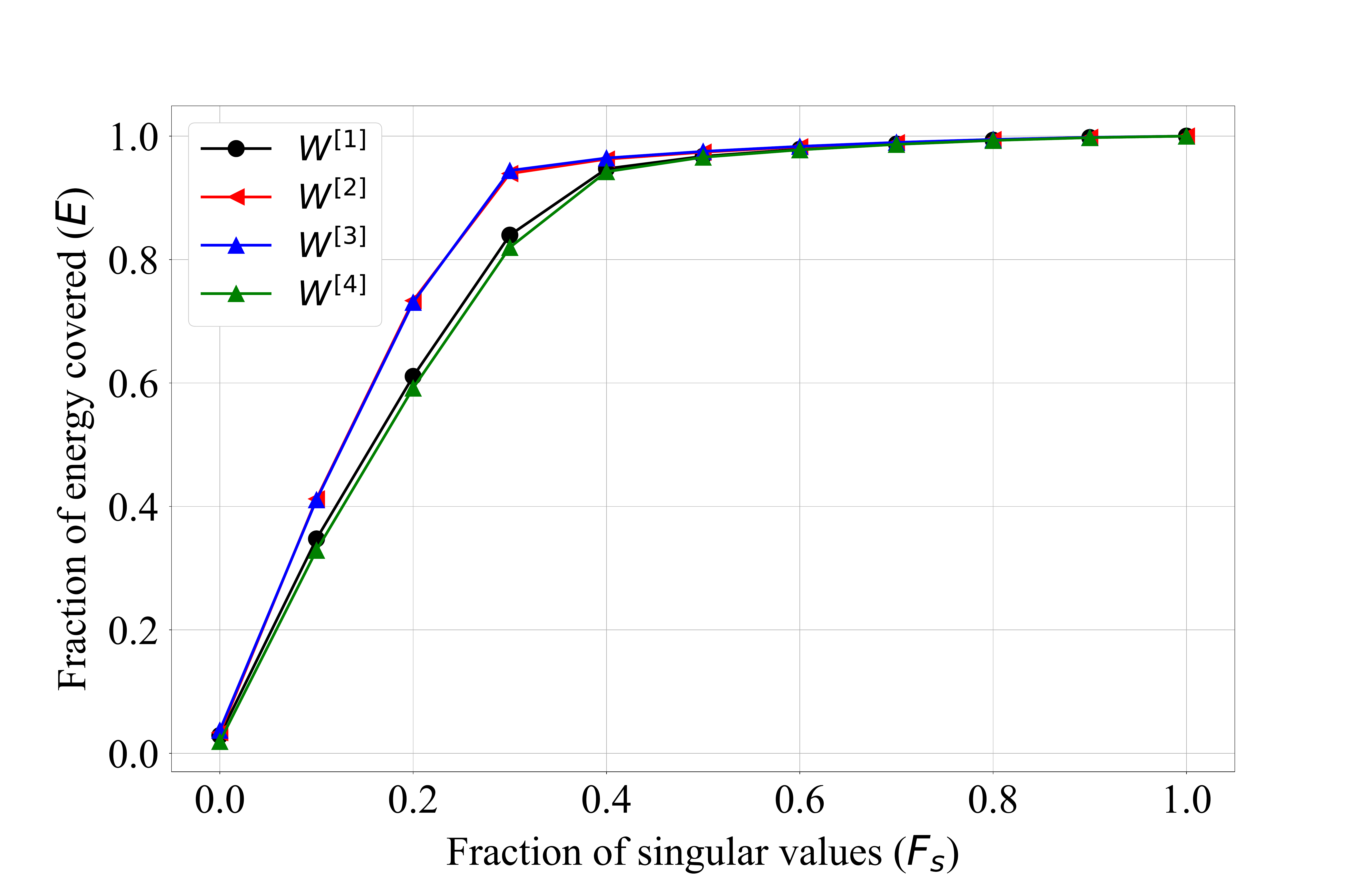}} \\
\subfloat[\label{fig:test2}]
  {\includegraphics[width=3.5in,height=1.8in]{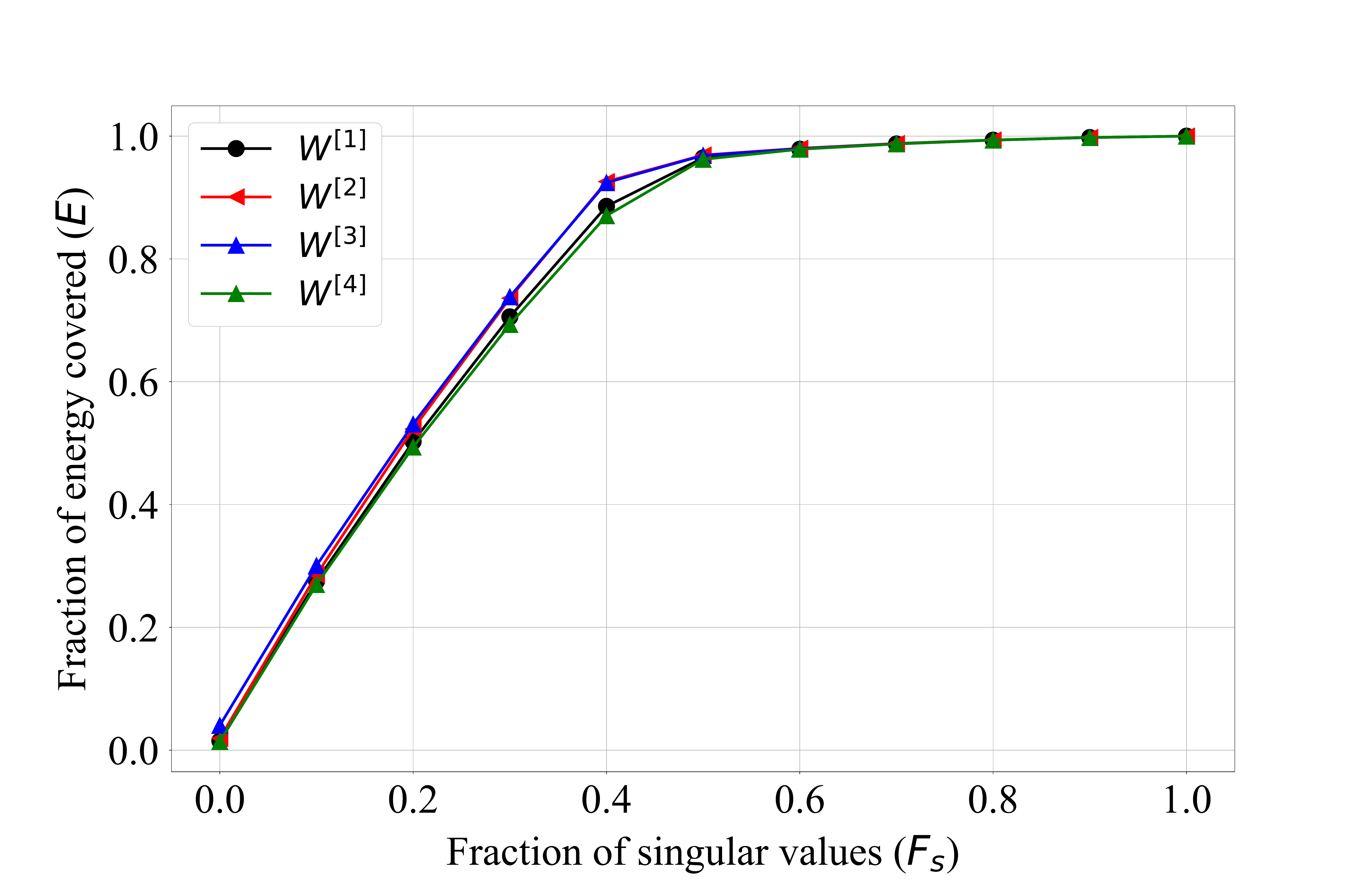}}\vfill
\caption{Energy distribution in weights matrices $W^{[l]}$ after training on (a) Keirn  (b) Motor EEG datasets using $algorithm 1$}
\label{x1}
\end{figure}
\begin{figure}[!ht]
\centering
\subfloat[\label{test1}]
  {\includegraphics[width=3.5in,height=1.8in]{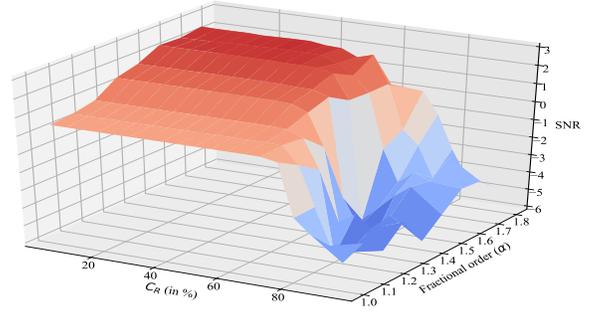}}\\
\subfloat[\label{fig:test2}]
  {\includegraphics[width=3.5in,height=1.8in]{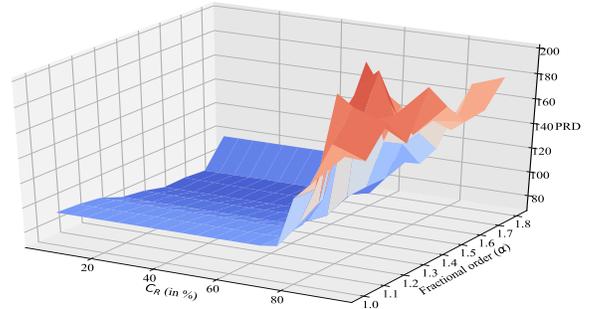}}\vfill
\caption{Variation of performance metrics (a) $SNR$ (b) $PRD$ with compression ratio $C_{R}$ and fractional order $\alpha$ using Keirn dataset}
\label{fg}
\end{figure}
\begin{figure}[!ht]
\centering
\subfloat[\label{test2}]
  {\includegraphics[width=3.5in,height=1.8in]{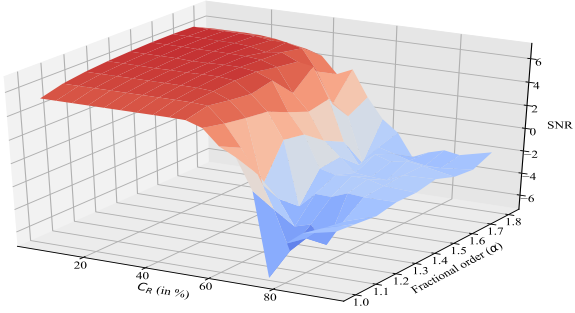}} \\
\subfloat[\label{fig:test2}]
  {\includegraphics[width=3.5in,height=1.8in]{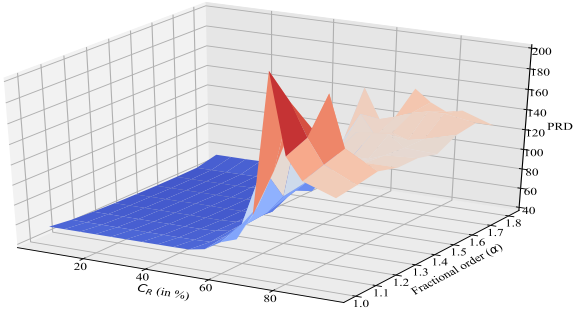}}\vfill
\caption{Variation of the performance metrics (a) $SNR$ (b) $PRD$ with compression ratio $C_{R}$ and fractional order $\alpha$ using Motor EEG dataset}
\label{fg1}
\end{figure}
\subsection{Compression analysis}
Deep learning architectures are both computationally and memory intensive, thereby making them difficult to deploy on real-time systems. Therefore, compressing the architectures makes them portable so that they can be used in edge computing devices such as mobiles. The compression technique should be such that the performance of the architecture is not degraded, too much when the weights are compressed. Now, the denoising performance of the proposed architecture as a function the weight compression $C_{R}$ is evaluated.\\
To compress the weight matrices optimally, we have to  ensure that the compressed weights $W_{c}^{[l]}$ contain most of the energy present in the original weights $W^{[l]}$. The plots in Figs. \ref{x1}(a)-(b) show the energy distribution $(E)$ present in the weight matrices $W^{[l]}$ as a function of singular values $(F_{s})$ for Keirn and Motor EEG dataset, respectively. The parameters $E$ and $F_{s}$ are defined as  
\begin{equation}
E = \frac{\sum\limits_{i=1}^{r}S_i}{\sum\limits_{i=1}^{R}S_i}
\label{}
\end{equation}
and
\begin{equation}
F_s = \frac{r}{R}
\label{}
\end{equation}
Here, $R$ is the original rank of a matrix and $S$ is the singular value obtained from the RSVD operation on the weight matrix mentioned in $Algorithm \; 3$. It can be observed from the figures that when $F_{s}$ is in the range of $(0.35- 0.4)$, $90\%$ of the energy $E$ present in the $W^{[l]}$ is preserved. Hence, by retaining a fewer number of singular values helps in obtaining the compressed weight matrices $W_{c}^{[l]}$. Further, it can be seen that the energy profiles for the pairs  ($W\textsuperscript{[2]}$,$W\textsuperscript{[3]}$) and ($W\textsuperscript{[1]}$,$W\textsuperscript{[4]}$) follow a similar pattern. This is due to the fact that the proposed architecture is a symmetrical auto-encoder in terms of the weight matrices.
\begin{figure}[!htbp]
\centering
\subfloat[\label{test3}]
  {\includegraphics[width=3.5in,height=1.8in]{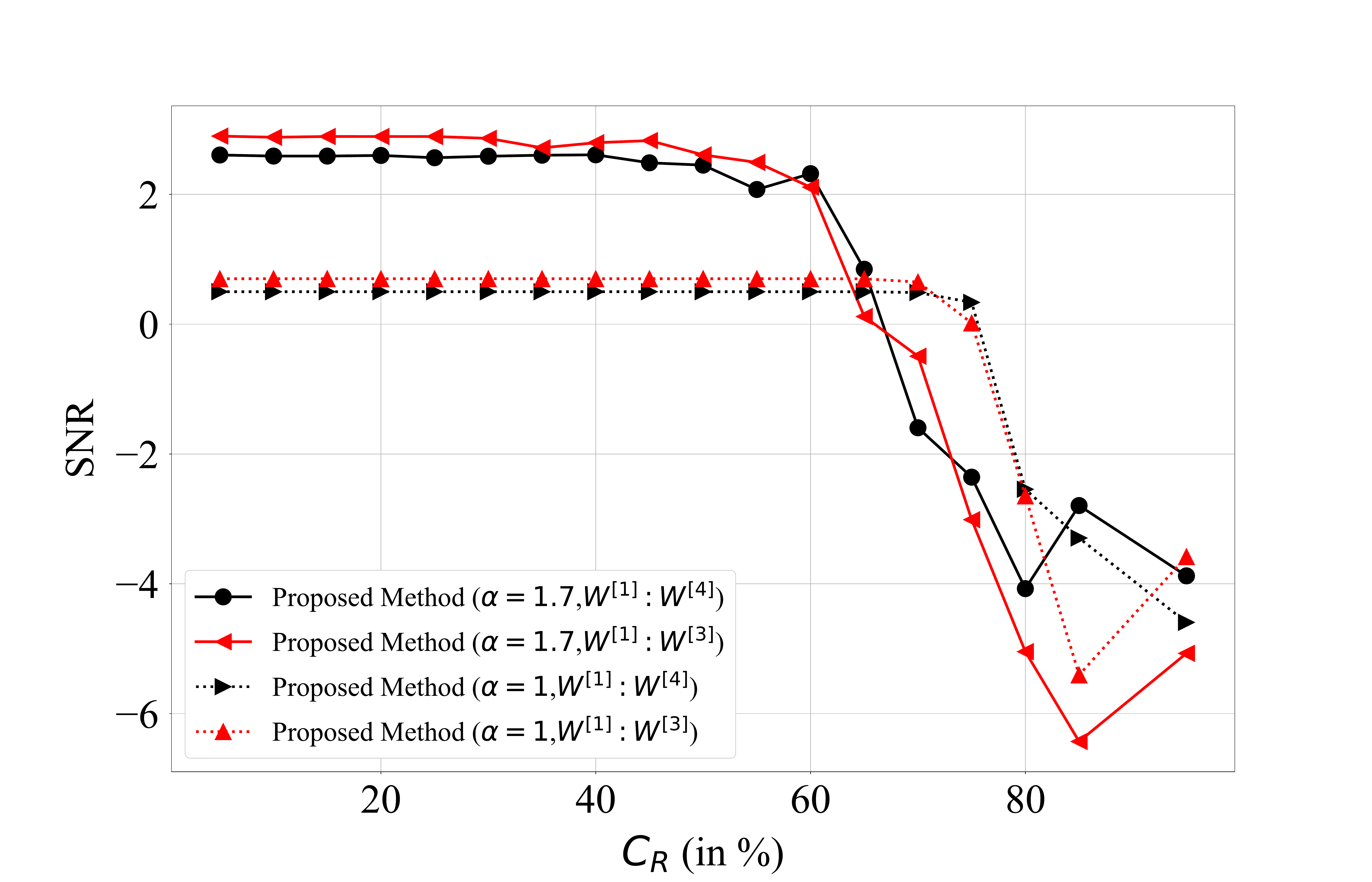}}\\
\subfloat[\label{fig:test2}]
  {\includegraphics[width=3.5in,height=1.8in]{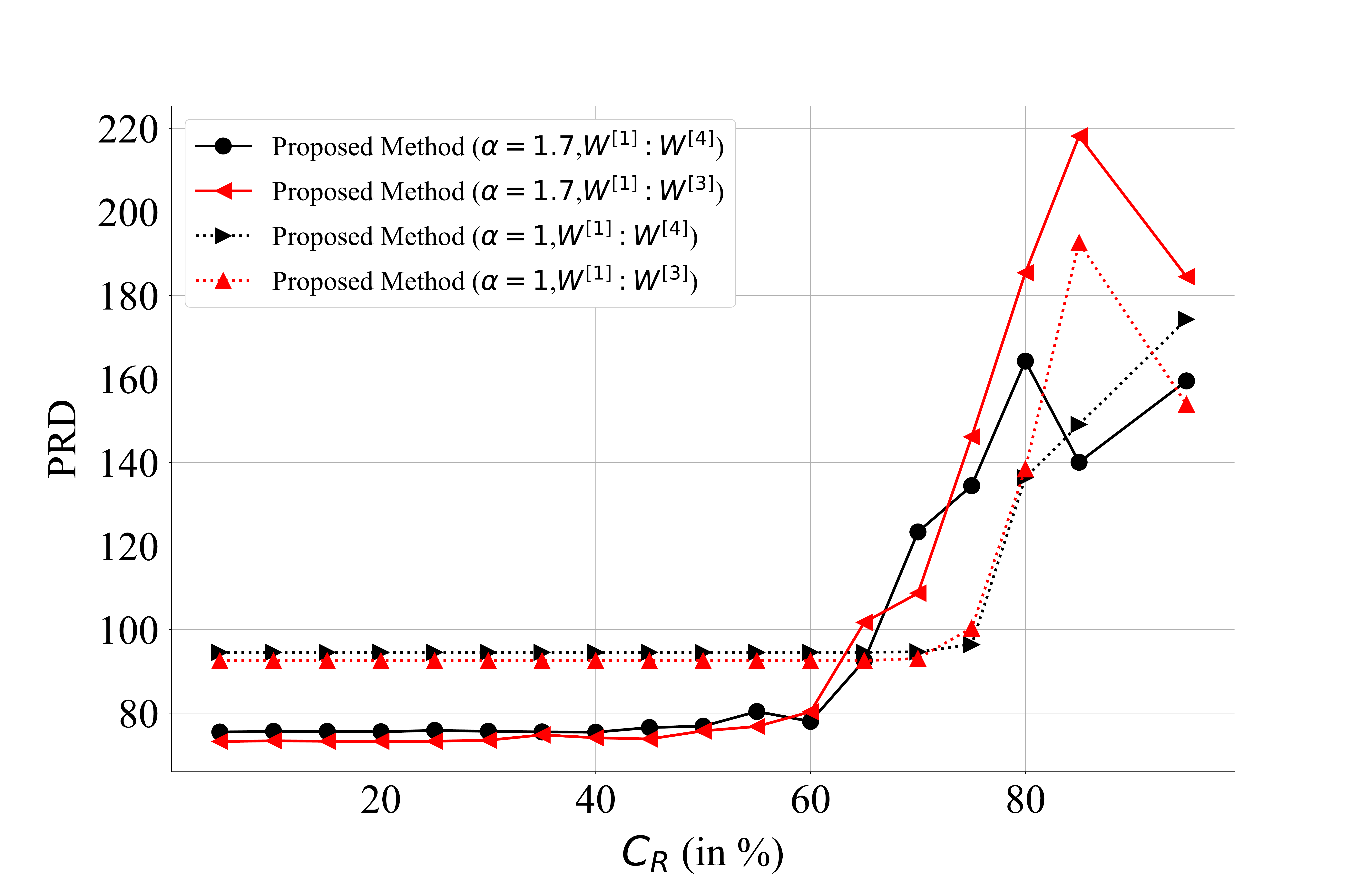}}\vfill
\caption{Sensitivity analysis of weight compression on performance metrics  (a) $SNR$ (b) $PRD$ for Keirn dataset }
\label{u1}
\end{figure}
\begin{figure}[!htbp]
\centering
\subfloat[\label{test3}]
  {\includegraphics[width=3.5in,height=1.8in]{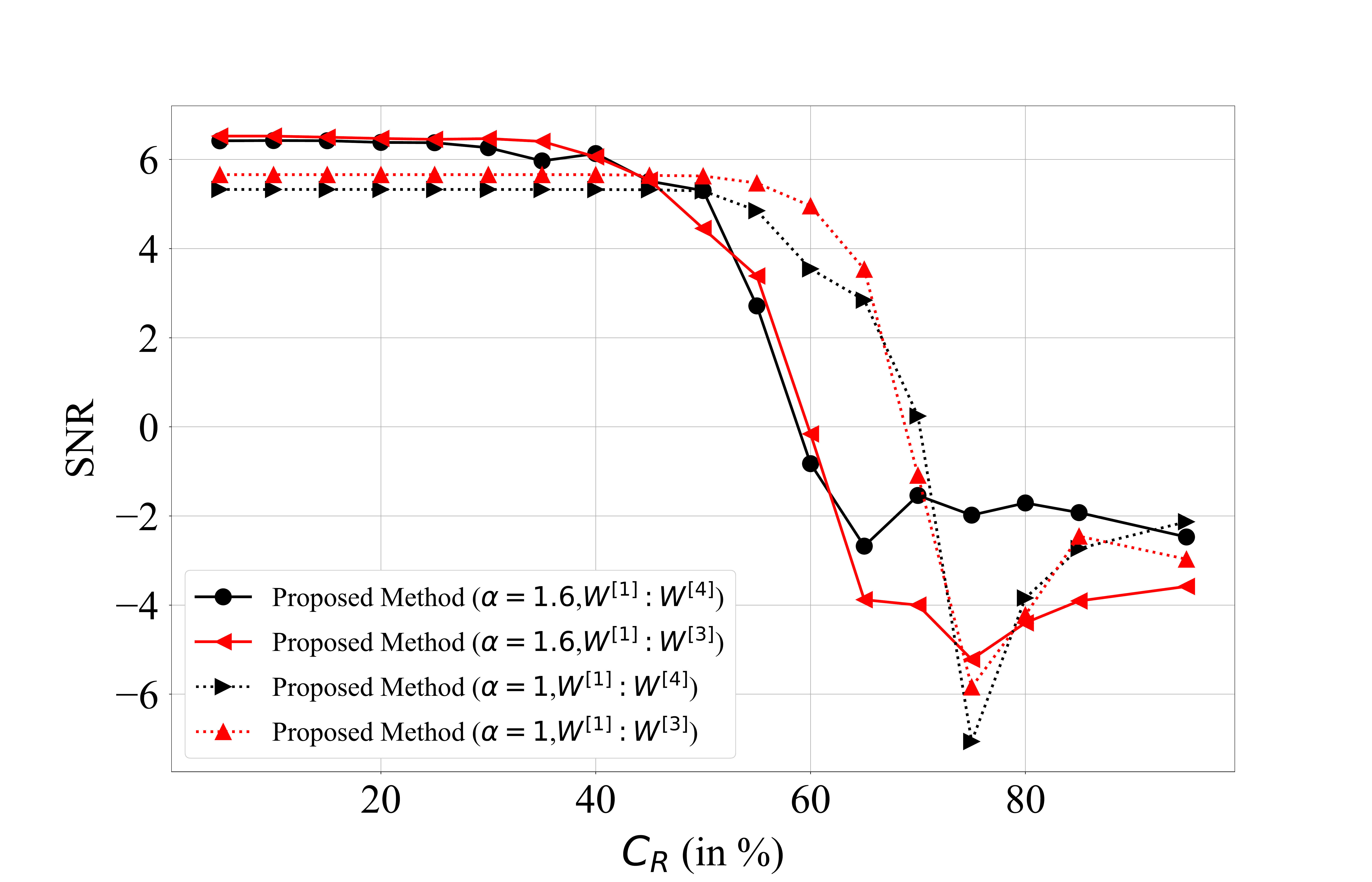}}\\
\subfloat[\label{fig:test2}]
  {\includegraphics[width=3.5in,height=1.8in]{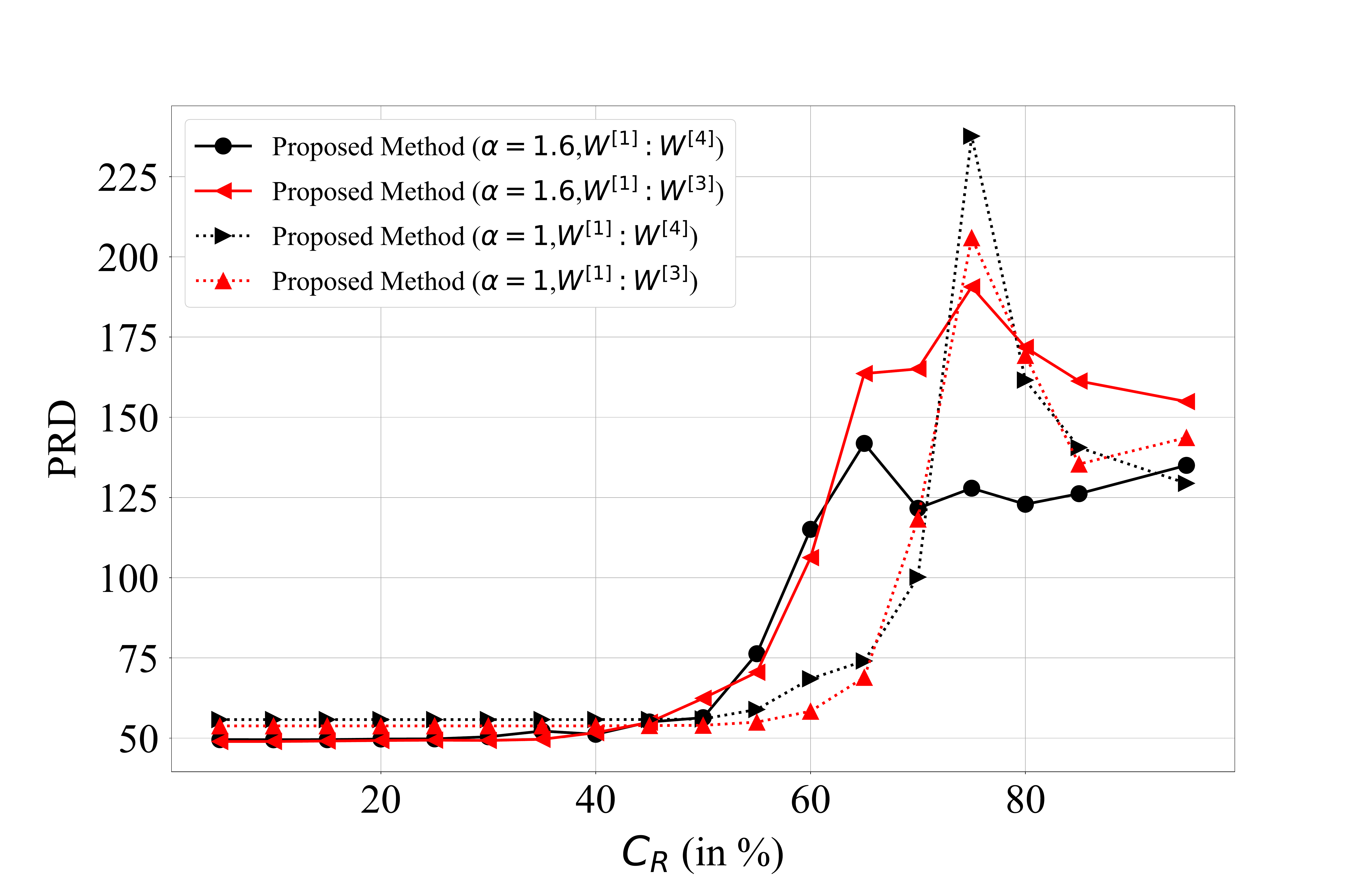}}\vfill
\caption{ Sensitivity analysis of weight compression on performance metrics  (a) $SNR$ (b) $PRD$ for Motor EEG dataset}
\label{u2}
\end{figure}
The performance of the auto-encoder is evaluated using $Algorithm\;4$, once the compressed weights $W_{c}^{[l]}$ are obtained using Fig. \ref{x1}. The $SNR$ and $PRD$ values for both the datasets are plotted in Figs. \ref{fg} and \ref{fg1} with respect to varying fractional order ($\alpha$) and compression ratio $(C_{R})$. Here, $\alpha$ is varied from $1.0$ to $1.8$ in steps of $0.1$. It can be observed that when the weight matrices of the model are compressed up to $(55-60)\%$, it performs significantly better than the existing methods do, without much drop in the performance. The plot also suggests that fractional orders derivatives can be used along with the compression technique (RSVD) to boost the  performance loss, which normally occurs if only compression is performed on conventional auto-encoders with $\alpha = 1$.\\
Sensitivity analysis of the performance metrics has been carried out are shown in Figs. \ref{u1} and \ref{u2} for different datasets. In this study, the variation of $SNR$ and $PRD$ values is recorded as a function of the weight compression performed on selective layers of auto-encoder. The $SNR$ and $PRD$ results plotted for Keirn dataset in Figs. \ref{u1}(a)-(b) indicate that compressing the first three weights ($W^{[1]}:W^{[3]}$) gives better performance upto a $C_{R}$ of $55\%$, and around $C_{R}$ of $(60-65)\%$, the model with all weights ($W^{[1]}:W^{[4]}$) compressed give better denoising results. Due to this reason, compression has been carried out on all the weights of the auto-encoder. Similarly, it can be observed from Figs. \ref{u2}(a)-(b) that for the Motor EEG dataset, at a $C_{R}$ of $(50-55)\%$ the fractional auto-encoder with all weights compressed gives performance similar to that, when compared to its first three weights compressed. Moreover, observing both the figures it is evident that if the $C_{R}$ is restricted between $(50-60)\%$, the denoising performance of the fractional auto-encoder ($\alpha = 1.7,1.6$) is better compared to its conventional counterpart ($\alpha = 1$).
\section{Conclusion} \label{sec4}
We have introduced a fractionally compressed auto-encoder for the removal of Gaussian noise from EEG signals. The auto-encoder uses Tchebichef moments as orthogonal features in the input. It has been shown that the proposed model performs better than the existing methods for denoising EEG signals. Furthermore, the RSVD compression is also applied on the auto-encoder weights to show that even after $55\%$ compression, the model performs well on fractional order of $1.6$ and $1.7$ for Motor EEG and Keirn datasets, respectively. Future work involves the deployment of the compressed architecture as an open source application so that it can be easily imported on edge computing devices.

\bibliographystyle{unsrt}
\bibliography{references.bib}
\end{document}